\documentclass[letterpaper, 10 pt, conference]{ieeeconf}

\IEEEoverridecommandlockouts
\usepackage{cite}
\usepackage{amsmath,amssymb,amsfonts}
\usepackage{algorithmic}
\usepackage{graphicx}
\usepackage{textcomp}
\usepackage{xcolor}
\usepackage{tabularx}
\usepackage{caption}
\usepackage{multirow}

\def\BibTeX{{\rm B\kern-.05em{\sc i\kern-.025em b}\kern-.08em
    T\kern-.1667em\lower.7ex\hbox{E}\kern-.125emX}}

\newcommand{\todo}[1]{{\color{blue}{#1}}}
\newcommand{\changed}[1]{{\color{red}{#1}}}

\begin{document}

\title{
\textbf{
Grasp, Slide, Roll: Comparative Analysis of Contact Modes for Tactile-Based Shape Reconstruction\\
}

\thanks{*This work was done when Chung Hee Kim and Shivani Kamtikar was an intern at Amazon.}
\thanks{
$^\ddagger$Ta\c{s}k{\i}n Pad{\i}r holds concurrent appointments as a Professor of Electrical and Computer Engineering at Northeastern University and as an Amazon Scholar. This paper describes work performed at Amazon and is not associated with Northeastern University.
}%
\thanks{$^{1}$Amazon Fulfillment Technologies \& Robotics, Westborough MA, USA
    {\tt\footnotesize \{bradytye, ptaskin, jmigdal\}@amazon.com}}%
\thanks{$^{2}$Robotics Institute at Carnegie Mellon University, Pittsburgh PA, USA \texttt{chunghek@andrew.cmu.edu}}
\thanks{$^{3}$Siebel School of Computing and Data Science, University of Illinois at Urbana-Champaign, Champaign IL, USA
        {\tt\footnotesize skk7@illinois.edu}}%

}

\author{Chung Hee Kim$^{1,2}$, Shivani Kamtikar$^{1,3}$, Tye Brady$^{1}$, Ta\c{s}k{\i}n Pad{\i}r$^{1}$, Joshua Migdal$^{1}$}

\maketitle

\begin{abstract}
Tactile sensing allows robots to gather detailed geometric information about objects through physical interaction, complementing vision-based approaches. However, efficiently acquiring useful tactile data remains challenging due to the time-consuming nature of physical contact and the need to strategically choose contact locations that maximize information gain while minimizing physical interactions. This paper studies how different contact modes affect object shape reconstruction using a tactile-enabled dexterous gripper. We compare three contact interaction modes: grasp-releasing, sliding induced by finger-grazing, and palm-rolling. These contact modes are combined with an information-theoretic exploration framework that guides subsequent sampling locations using a shape completion model. Our results show that the improved tactile sensing efficiency of finger-grazing and palm-rolling translates into faster convergence in shape reconstruction, requiring 34\% fewer physical interactions while improving reconstruction accuracy by 55\%. We validate our approach using a UR5e robot arm equipped with an Inspire-Robots Dexterous Hand, showing robust performance across primitive object geometries.
\end{abstract}

\vspace{-3pt}
\section{Introduction}

Robotic manipulation systems have traditionally relied on vision-based sensing to guide their interactions with objects. While effective in controlled environments, these systems often struggle when confronted with dynamic scenarios where objects shift during manipulation or in cluttered environments where visual occlusion becomes a significant challenge. These limitations stem from their dependence on pre-touch sensing and inability to adapt to real-time changes in the environment. Tactile sensing emerges as a promising complement to visual perception, offering robots the ability to gather precise geometric information through direct physical contact, much like the human sense of touch.

\begin{figure}[t]
\centering 
\includegraphics[width=\columnwidth]{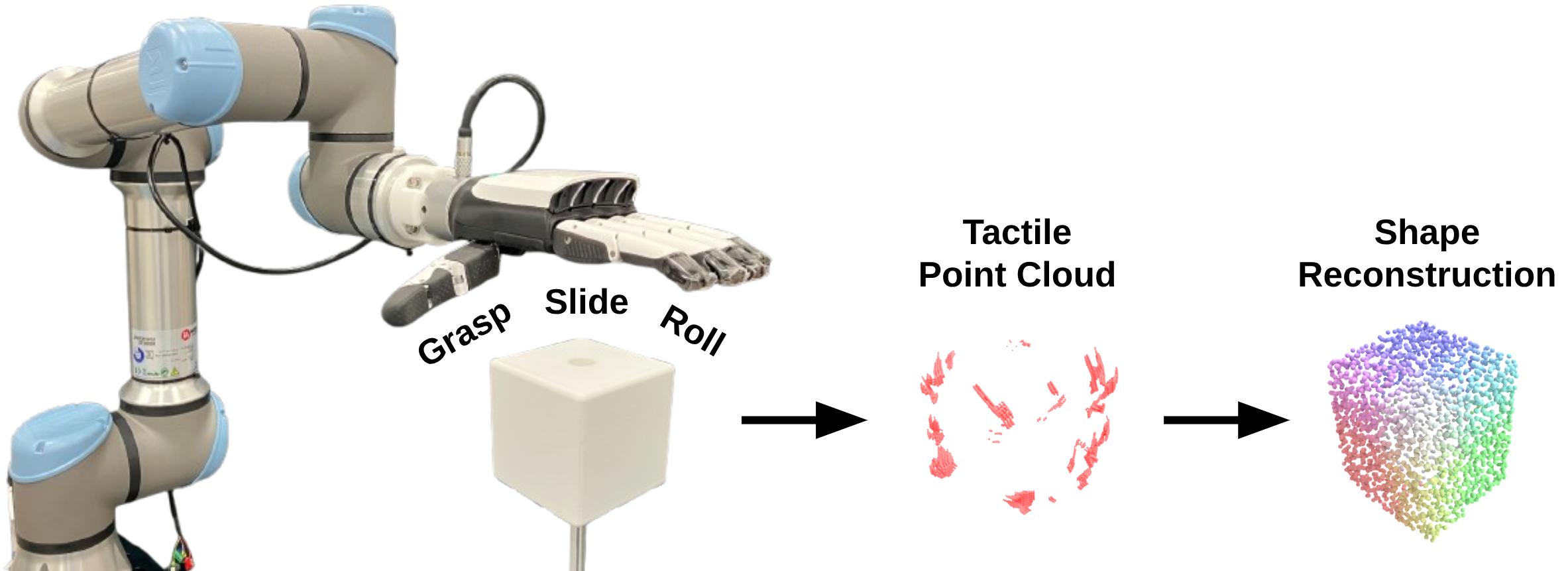}
\setlength{\unitlength}{1cm}
\begin{picture}(0,0)
\end{picture}
\caption{Tactile information is promising to unlock contact-rich manipulation capabilities for robots. 
We investigate tactile data acquisition efficiency of various contact modes (sliding, rolling, and discrete contacts) in the context of 3D object reconstruction using tactile data only (no vision).}
\label{fig:main_figure} 
\vspace{-10pt}
\end{figure}

However, efficiently acquiring and utilizing tactile data presents several key challenges. Naive interaction strategies, such as simple grasp-and-release actions, often yield limited contact information, making accurate object reconstruction slow and impractical in real-world settings. Furthermore, the scarcity of large-scale tactile datasets limits the development of robust tactile perception algorithms. Together, these challenges highlight a central trade-off in tactile perception: richer object geometry estimates require more frequent contacts, yet practical systems must limit interactions to remain efficient. This motivates our central research question: \textit{How can minimal tactile interaction be leveraged to enable reliable, efficient object identification and maximize geometric reconstruction accuracy?}

This paper presents a novel methodology to explore contact-rich object interactions for dexterous manipulators with tactile sensing through three main contributions. First, we employ a diffusion-based shape completion model that can reconstruct complete object geometry from sparse tactile measurements with zero-shot sim-to-real transfer. Second, we develop an information-theoretic exploration framework that strategically guides tactile sampling to maximize information gain. Thirdly, we present a comprehensive study of different contact interaction modes, comparing the efficiency of \textit{grasp-releasing}, \textit{finger-grazing}, and \textit{palm-rolling} in tactile data acquisition for object reconstruction, characterized from contact kinematics of two surfaces \cite{1087971}. 

We demonstrate our approach on a real robotic system comprising an anthropomorphic dexterous hand equipped with distributed tactile sensor arrays mounted on a 6-DOF robotic arm (see Fig.~\ref{fig:main_figure}). Our experimental results show that dynamic contact modes (sliding and rolling) guided by our information-theoretic exploration policy achieve better performance compared to basic grasp-and-release interactions, requiring 34\% fewer contacts while improving reconstruction accuracy by 55\%. The system demonstrates robust shape completion across various object geometries, requiring an average of 8.4 interactions per object for accurate reconstruction. These findings highlight the critical role of diverse contact modes in robotic tactile exploration. Going beyond simple grasping can enhance tactile information acquisition, motivating exploration frameworks that strategically select interaction modes for richer object understanding.

\section{Related Work}
\label{sec:literature}
\subsection{Tactile Sensing for Robotic Manipulation}

Tactile sensing has been gaining attention as a rich feedback modality for enhancing robotic dexterity. Model-based approaches leverage tactile data to estimate object properties such as friction coefficients \cite{8453829}, slippage \cite{5509330}, material characteristics \cite{9811543}, as well as object pose estimation using tactile sensing integrated with vision \cite{9709520, li2025}. Grounded in physical principles, these methods offer valuable insights into tactile sensing; however, they often assume simplified gripper designs or rely on analytical models that may not generalize across diverse contact conditions and nonlinear interactions \cite{kappassov:hal-01680649}. Recent trends shift toward deep learning-based approaches that directly utilize high-dimensional tactile data from vision-based tactile sensors for in-hand manipulation \cite{9018215, guzey2023dexterity, 7363524, khandate2023sampling} and tactile servoing \cite{8794219}. While these approaches remove the need for explicit modeling, they tend to struggle with out-of-distribution generalization and lack structured strategies for contact planning.

\subsection{Tactile-Based 3D Reconstruction}

A growing body of research investigates tactile-guided shape reconstruction, often in combination with vision \cite{10720798}. Early methods fuse depth images with tactile contacts for voxel-based reconstruction \cite{8593430}, while more recent approaches employ neural fields for in-hand visuotactile SLAM \cite{doi:10.1126/scirobotics.adl0628} or actively select contact regions based on visual priors \cite{smith2021active, oikonomou2025proactive}. Exploration strategies vary from learned tactile-based policies \cite{wang2025touch2shapetouchconditioned3ddiffusion, 10611667} to random sampling \cite{comi2024touchsdf} and heuristic approaches \cite{LUAN2025117350, 10766628}. With the emergence of 3D Gaussian Splatting, several studies have investigated vision-tactile fusion for high-fidelity surface reconstruction \cite{10802412, comi2025snapit, fang2025fusionsense}. While these methods demonstrate improved surface modeling, they primarily treat tactile sensing as a complementary cue to vision. In contrast, our work focuses on diverse tactile interaction strategies to enable complete object reconstruction from purely tactile feedback, without reliance on visual input.

\subsection{Tactile-Sensing Anthropomorphic Grippers}

Several anthropomorphic end-effectors have been developed with embedded tactile sensors \cite{5152650, 5771603, 4058994}, yet many are limited to hardware demonstrations without control or perception strategies. Some works integrate tactile sensing with Bayesian and RL-based exploratory behaviors for object identification \cite{6631001} or implement force control loops for manipulation \cite{chelly2025tactilebasedforceestimationinteraction}. However, most of these efforts do not explicitly model or analyze different modes of contact. Reconstruction methods using tactile-equipped anthropomorphic hands have shown promising results \cite{doi:10.1126/scirobotics.adl0628, smith2021active}, but none investigate the role of distinct contact modes in shaping object understanding. Our work is the first to study the impact of contact modes on tactile shape exploration using an anthropomorphic, tactile-enabled hand, offering new insights into how diverse exploratory behaviors can be harnessed for efficient reconstruction.

\section{Sparse Tactile Sensing to Dense Geometry}
\label{sec:spvd}

This section presents our approach to reconstructing complete object geometry from sparse tactile measurements, detailing the shape completion model, key improvements for real-world deployment, and the synthetic data generation pipeline.

\subsection{Diffusion-Based 3D Shape Completion}
\label{subsec:completion_model}

For 3D shape completion, we build upon and enhance the Sparse Point-Voxel Diffusion (SPVD) model \cite{romanelis2024efficient}, with key improvements enabling position, scale, and rotation invariance 
(detailed in Sec.~\ref{subsec:enhancements}). Specifically designed to diffuse point clouds, SPVD's dual-space architecture combines point-space processing for local feature extraction with voxel-space encoding for global shape understanding through 3D convolutions. The model implements a diffusion process that iteratively denoises $N$ randomly sampled points $P_{pred}=\{p_i\in \mathbb{R}^3 \mid i = 1,...,N\}$, conditioned on tactile measurements represented by a partial point cloud $P_{meas}=\{p_j\in \mathbb{R}^3 \mid j= 1,...,M\}$. Starting from Gaussian noise, the model progressively refines $P_{pred}$ through $T$ denoising steps to generate a dense point cloud representing the complete object geometry (see Fig.~\ref{fig:synthetic_results}). 

\subsection{Improvements for Real-World Deployment}
\label{subsec:enhancements}
The original SPVD model was designed for clean 3D assets, making it unsuitable for direct use with tactile data. Real-world tactile sensing presents several challenges absent in simulations. Physical sensor readings contain measurement noise and drift, objects are encountered in random poses, and their scale can vary arbitrarily. Additionally, the reference frame for tactile observations depend on the robot's configuration, unlike standard 3D shape assets where canonical object poses are readily available. 
To enable real-world tactile shape completion, we introduced critical enhancements and data augmentations that were necessary for the model to process tactile measurements:

\subsubsection{Model Enhancements} 
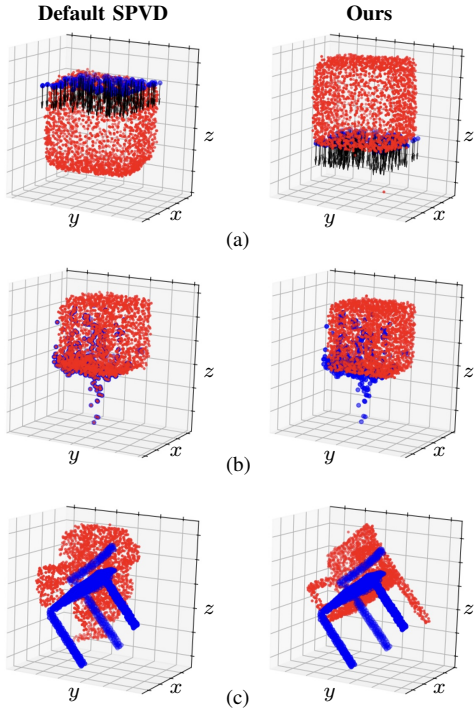
\begin{figure}[t]
\vspace{10pt}
\centering 
\includegraphics[width=0.75\columnwidth]{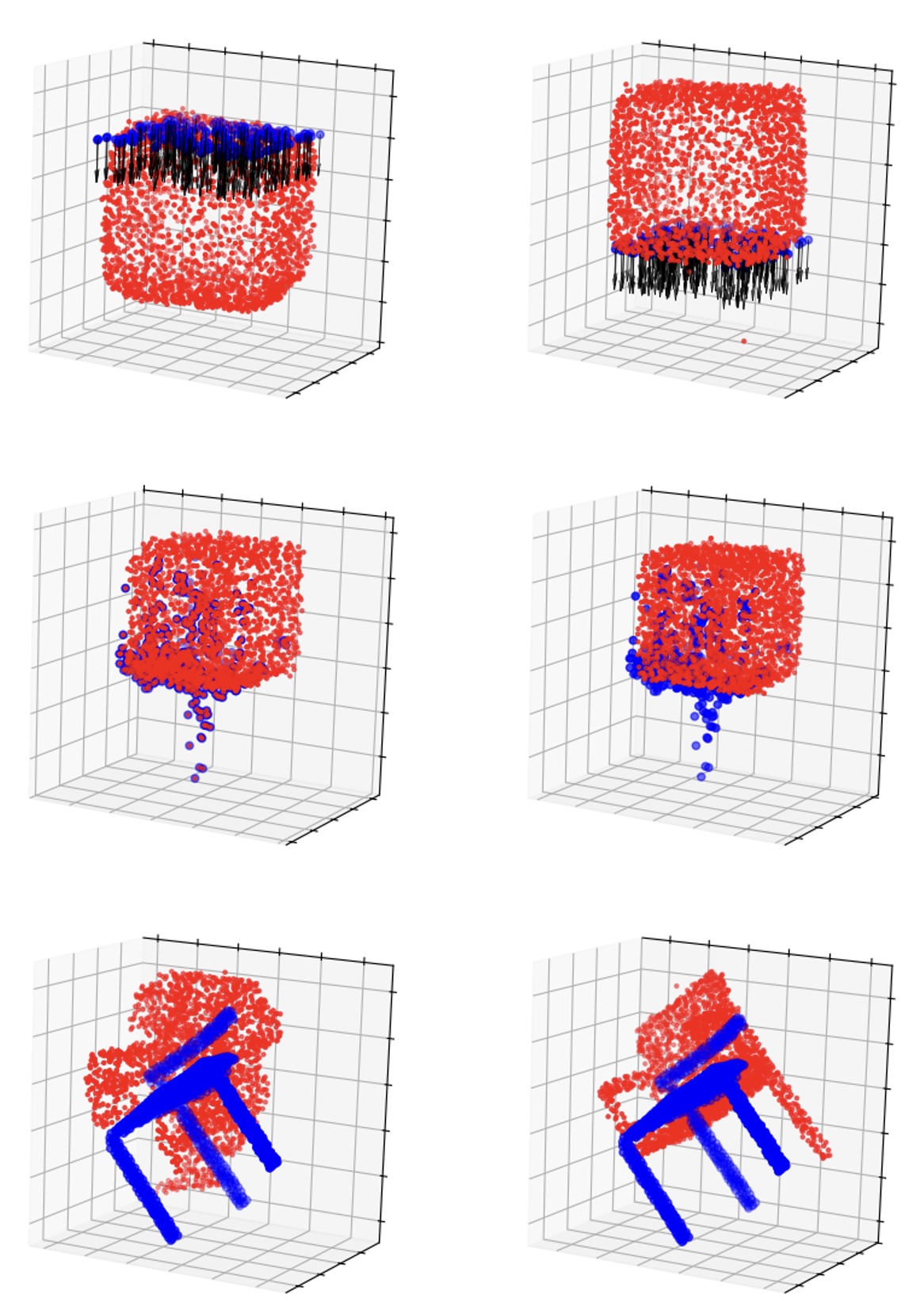}
\setlength{\unitlength}{1cm}
\begin{picture}(0,0)
\put(-6.1, 9.1){\footnotesize \textbf{Default SPVD}}
\put(-2.0, 9.1){\footnotesize \textbf{Ours}}

\put(-5.7, 6.4){\footnotesize $y$}
\put(-4.35, 6.45){\footnotesize $x$}
\put(-3.9, 7.5){\footnotesize $z$}

\put(-2.3, 6.4){\footnotesize $y$}
\put(-0.85, 6.45){\footnotesize $x$}
\put(-0.45, 7.5){\footnotesize $z$}

\put(-5.7, 3.25){\footnotesize $y$}
\put(-4.35, 3.3){\footnotesize $x$}
\put(-3.9, 4.35){\footnotesize $z$}

\put(-2.3, 3.25){\footnotesize $y$}
\put(-0.85, 3.3){\footnotesize $x$}
\put(-0.45, 4.35){\footnotesize $z$}

\put(-5.7, 0.1){\footnotesize $y$}
\put(-4.35, 0.15){\footnotesize $x$}
\put(-3.9, 1.2){\footnotesize $z$}

\put(-2.3, 0.1){\footnotesize $y$}
\put(-0.85, 0.15){\footnotesize $x$}
\put(-0.45, 1.2){\footnotesize $z$}

\put(-3.6, 6.1){\footnotesize (a)}
\put(-3.6, 3.1){\footnotesize (b)}
\put(-3.6, 0.0){\footnotesize (c)}
\end{picture}

\caption{Comparative examples of original SPVD results and our improved approach. Partial input points and shape completed output points are colored blue and red, respectively. (a) Shape completion with (right) and without (left) \textbf{normal vectors} (shown as black arrows) illustrates how normal guidance disambiguate shape generation direction. (b) \textbf{Novel point generation} filters noisy measurements (right), while the original approach preserves noisy inputs (left). (c) \textbf{Rotation-invariant} shape completion achieved through random rotation augmentation, demonstrated with a chair point cloud.}
\label{fig:comparison} 
\end{figure}

We applied three key modifications to enable the model's real-world applicability:
\begin{itemize}
    \item \textbf{Point Normal Input Features:} 
    We enrich the input representation by incorporating surface normal vectors at each point. These normal vectors, which can readily be approximated as perpendicular to the contact surface, provide directional guidance for shape generation. Fig.~\ref{fig:comparison}(a) demonstrates how the absence of contact normal vectors leads to incorrect shape completion (left), while incorporating normal information enables reconstruction aligned with the true surface orientation (right).
    \item \textbf{Novel Point Generation:} We modified the model to generate entirely new output points, departing from SPVD's default approach where measured points are retained in the final prediction ($P_{meas} \subseteq P_{pred}$). While the assumption that input points belong to the true shape works for clean data, tactile measurements often contain noise. Our modification ensures the predicted point cloud is distinct from the input ($P_{meas} \cap P_{pred} = \emptyset$), enabling the model to effectively filter noisy inputs and handle input point sets of unconstrained size. Fig.~\ref{fig:comparison}(b) illustrates this noise-filtering capability.
    \item \textbf{Position and Scale Invariance:} We applied per-sample normalization to achieve position and scale invariance. For each input point cloud $P=P_{meas}$, we apply:
    \begin{equation}
        P'=(P-\bar{P})
    \end{equation}
    \begin{equation}
        P_{normalized} = P'/ (\lambda\sigma_{||P'||_2})
    \end{equation}
where $\bar{P}$ represents the mean, $\lambda$ is a scaling factor, and $\sigma_{||P'||_2}$ denotes the standard deviation of L2-norms of the centered points $P'$. This normalization enables origin-independent and scale invariant predictions during inference.
\end{itemize}

\subsubsection{Data Augmentation} 
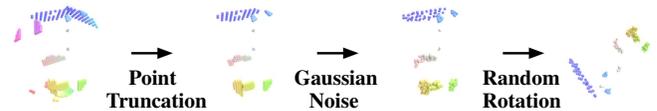
\begin{figure}[t]
\centering 
\includegraphics[width=\columnwidth]{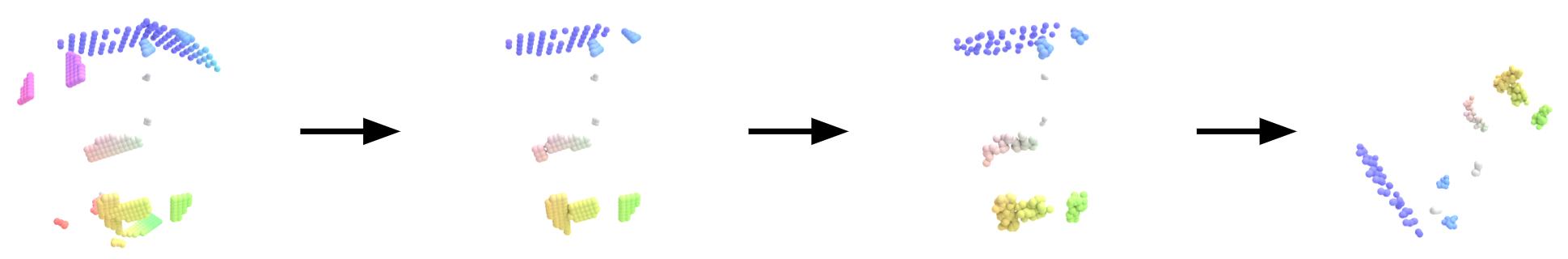}
\setlength{\unitlength}{1cm}
\begin{picture}(0,0)
\put(-2.7, 0.7){\footnotesize \textbf{Point}}
\put(-3.0, 0.4){\footnotesize \textbf{Truncation}}
\put(-0.5, 0.7){\footnotesize \textbf{Gaussian}}
\put(-0.3, 0.4){\footnotesize \textbf{Noise}}
\put(2.0, 0.7){\footnotesize \textbf{Random}}
\put(2.0, 0.4){\footnotesize \textbf{Rotation}}
\end{picture}
\caption{Data augmentation pipeline on a simulated tactile point cloud including contact point truncation, Gaussian noise injection, and random rotation, illustrated using contact data from a sphere. Points are colored to show correspondence before and after applying the augmentation.}
\label{fig:augmentation} 
\end{figure}

To improve sim-to-real transfer and enhance model robustness, we applied the following data augmentations (illustrated in Fig.~\ref{fig:augmentation}):

\begin{itemize} 
    \item \textbf{Contact Point Truncation:} Number of input points were varied by truncating $P_{meas}$ in reverse chronological order, simulating different contact sequences and partial object interactions.
    \item \textbf{Sensor Noise Simulation:} Gaussian noise was injected into $P_{meas}$ to emulate physical sensor noise.
    \item \textbf{Random Rotation:} Input/target point clouds were rotated arbitrarily during training, ensuring the model learns rotation-invariant representations of object geometry as illustrated in Fig.~\ref{fig:comparison}(c).
\end{itemize}

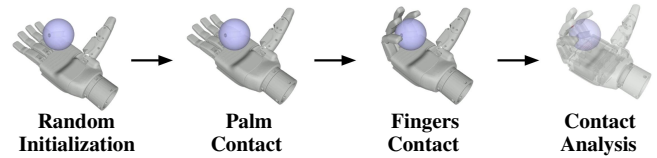
\begin{figure}[t]
\centering 
\includegraphics[width=\columnwidth]{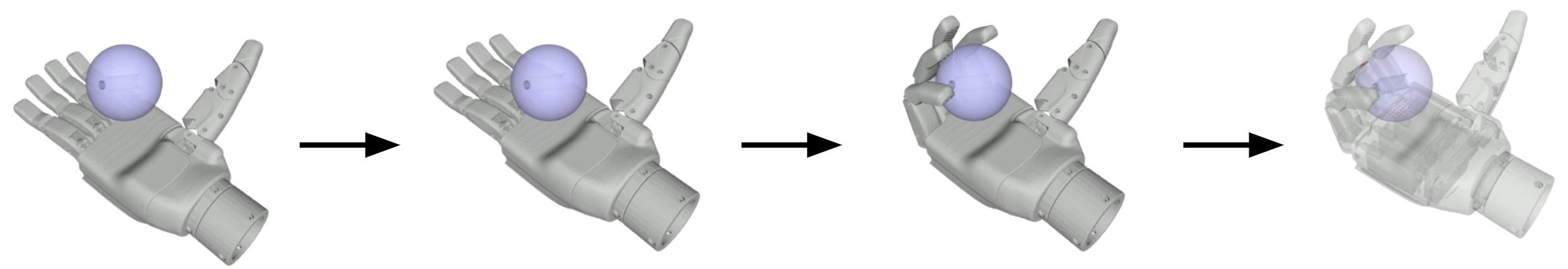}
\setlength{\unitlength}{1cm}
\begin{picture}(0,0)
\put(-3.85, 0.2){\footnotesize \textbf{Random}}
\put(-4.1, -0.1){\footnotesize \textbf{Initialization}}
\put(-1.35, 0.2){\footnotesize \textbf{Palm}}
\put(-1.55, -0.1){\footnotesize \textbf{Contact}}
\put(0.85, 0.2){\footnotesize \textbf{Fingers}}
\put(0.8, -0.1){\footnotesize \textbf{Contact}}
\put(3.15, 0.2){\footnotesize \textbf{Contact}}
\put(3.1, -0.1){\footnotesize \textbf{Analysis}}
\end{picture}
\vspace{4pt}
\caption{Contact motion sequence of the tactile contact simulator. The target object is a sphere shaded in purple.}
\label{fig:contact_simulator} 
\end{figure}

\subsection{Synthetic Tactile Data Generation}
To train the shape completion model, we developed a custom tactile contact simulator that generates synthetic training data using gripper forward kinematics and collision meshes (see Sec.~\ref{subsec:experiment_setup} for gripper specifications). The simulator emulates a four-step heuristic contact motion sequence (illustrated in Fig.~\ref{fig:contact_simulator}):
\begin{enumerate}
    \item The gripper is randomly initialized at a preset distance from the target object, with the palm facing the object.
    \item The gripper approaches the target object along the palm-normal axis until initial contact is detected.
    \item Each finger closes independently until object-finger contact is detected.  
    \item The positions of tactile contact points are computed through gripper-object mesh intersection analysis. 
\end{enumerate}
This process is repeated multiple times per object, resulting in a synthetic training dataset of paired tactile contact point clouds ($P_{meas}$) and ground truth object meshes, enabling efficient scaling of our training data.

\section{Tactile Exploration Framework}
This section presents our approach to address two critical challenges in tactile-driven perception: autonomous exploration and efficient data acquisition. We propose an information-theoretic framework for guiding optimal sampling locations, while investigating alternative contact interaction modes to improve tactile sensing efficiency.

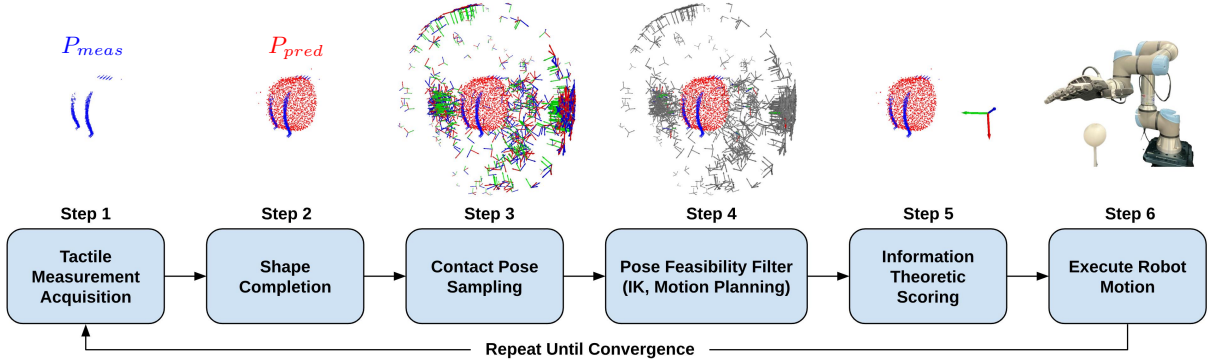
\begin{figure*}[t]
\centering 
\includegraphics[width=0.9\textwidth]{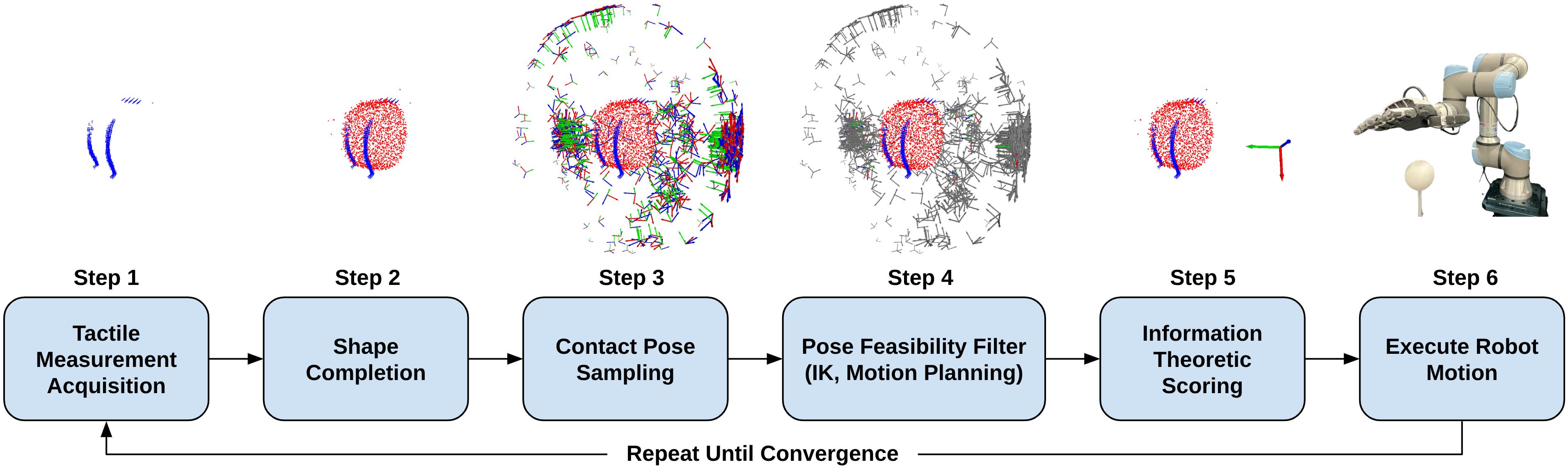}
\setlength{\unitlength}{1cm}
\begin{picture}(0,0)
\put(-15.5, 4.1){\small \todo{$P_{meas}$}}
\put(-12.75, 4.1){\small \changed{$P_{pred}$}}
\end{picture}
\caption{Process diagram of the information-theoretic tactile exploration. Measured tactile points $P_{meas}$ are colored blue (Step 1), while the shape completed point cloud $P_{pred}$ is colored red (Step 2). The coordinate frames in Step 3 represent candidate contact poses, while gray coordinate frames in Step 4 represents filtered candidates. The contact pose with the highest score is selected (Step 5) and executed (Step 6).}
\label{fig:pipeline} 
\end{figure*}

\subsection{Information-Theoretic Exploration} 
\label{subsec:exploration}
We present an exploration strategy that optimizes tactile sampling for 3D object reconstruction. Our framework iteratively estimates object geometry with a pretrained shape completion network based on tactile measurements. By quantifying uncertainties in these estimates, we compute information gain metrics to identify the most informative regions for subsequent sampling. The robot targets high-uncertainty areas, ensuring new tactile contacts improve our shape understanding. The framework illustrated in Fig.~\ref{fig:pipeline} enables systematic object exploration with the objective of shape reconstruction with minimal tactile interactions. 

\subsubsection{Tactile Measurement Acquisition} We obtain 3D tactile measurements through forward kinematics of the robot arm and hand. The tactile sensor locations are transformed from the hand frame to the base frame: 
\begin{equation} 
    P^{base}=T^{base}_{TCP}\space T^{TCP}_{hand} \space P^{hand} 
\end{equation} 
where $P^{base}$ and $P^{hand}$ represent the tactile point clouds in their respective frames, and the transformation matrices map between the robot base, tool center point (TCP), and hand frames.

\subsubsection{Shape Completion} A trained shape completion model (discussed in Sec.~\ref{sec:spvd}) processes the tactile measurements $P_{meas}$ to predict complete object geometry, establishing our current belief state $P_{pred}$ of the object shape.

\subsubsection{Contact Pose Sampling} We generate candidate contact poses based on both measured tactile points $P_{meas}$ and the predicted shape point cloud $P_{pred}$. For each point $p_i \in P_{pred}$, we compute a coverage distance: 
\begin{equation} 
    d_{min}(p_i)=\min_j || p_i - p_j||_2 \text{ where } p_j \in P_{meas}
\end{equation} 
which is the Euclidean distance to its nearest measured tactile point. Points exceeding a 5mm coverage threshold are importance-sampled ($k=500$) to identify underexplored regions. To convert these points into actionable 6-DoF end-effector poses $\mathbf{p}$, we align the hand's palm with the surface normal at each sampled point, oriented toward the object, with a uniformly sampled yaw angle as depicted in Fig.~\ref{fig:pipeline}. 

\subsubsection{Feasibility Filtering} Candidate poses undergo two-stage filtering: (1) inverse kinematics and joint limit checks, and (2) collision-free motion planning via cuRobo \cite{sundaralingam2023curobo}. We retain poses with valid robot trajectories $\mathcal{T}_\mathbf{p}$.

\subsubsection{Information-Theoretic Scoring}
We evaluate candidate poses using an information-gain versus motion-cost metric: 
\begin{equation}
    Score(\mathbf{p}) = IG_{coverage}(\mathbf{p}) - C_{motion}(\mathcal{T}_\mathbf{p})
\end{equation}

\noindent The information gain $IG_{coverage}$ measures the potential value of unexplored regions. We apply a Gaussian kernel mapping to convert the coverage distance $d_{min}(p_i)$ to a heuristic probability $P(p_i)$:
\begin{equation}
    P(p_i)=\exp(-\frac{d_{min}(p_i)^2}{2\sigma^2})
\end{equation}
Applying Shannon's definition of self-information \cite{6773024}:
\begin{equation}
    IG_{coverage}(p_i) = -log P(p_i)= \frac{d_{min}(p_i)^2}{2\sigma^2}
\end{equation}
where $\sigma$ is a bandwidth parameter controlling the sensitivity. This maps large (small) distances to greater (smaller) information gain, consistent with the intuition that predicted surfaces further from measured points represent regions of higher uncertainty.

The motion cost $C_{motion}$ is calculated as a weighted sum of joint-space movements along the planned trajectory $\mathcal{T}_\mathbf{p}$:
\begin{equation}
    C_{motion}(\mathcal{T}_\mathbf{p}) = \sum_{t=1}^{T} \sum_{j=1}^{6} w_j \left| q_{t,j} - q_{t-1,j} \right|
\end{equation}
where $q_{t,j}$ represents the position of joint $j$ at timestep $t$, and $w_j$ denotes joint-specific weights that allow for differential penalization of joint movements. The equation computes the total motion cost by summing weighted joint displacements across all timesteps.

\subsubsection{Execution} The robot executes the highest-scoring feasible trajectory and interacts with the object. The process repeats until shape prediction converges, identified when the Chamfer distance between $K$ consecutive beliefs remain below a threshold.

\begin{figure}[htbp]
\centering 
\includegraphics[width=\columnwidth]{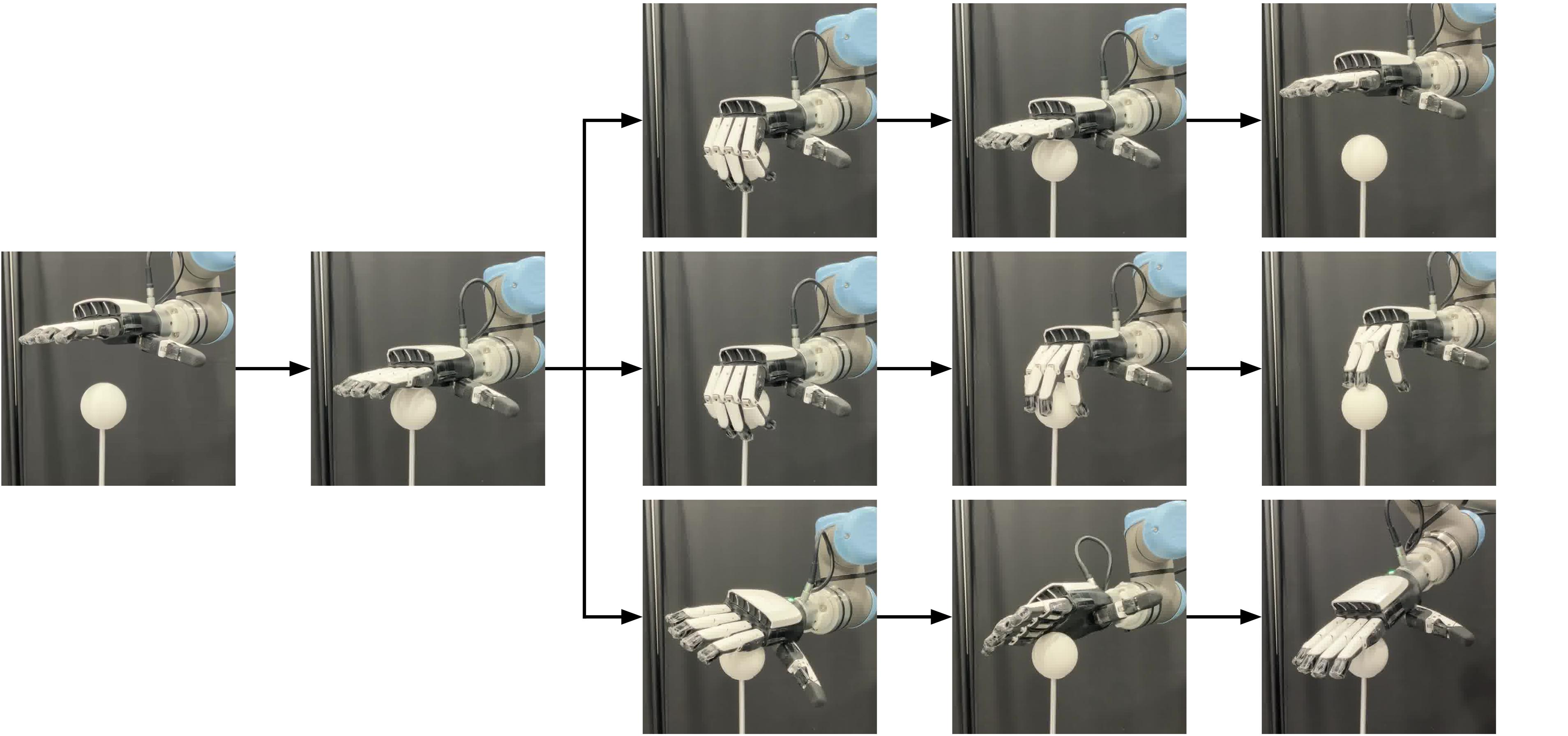}
\setlength{\unitlength}{1cm}
\begin{picture}(0,0)
\put(-4.05, 3.4){\scriptsize \textbf{Contact}}
\put(-3.88, 3.15){\scriptsize \textbf{Pose}}

\put(-2.25, 3.4){\scriptsize \textbf{Initial}}
\put(-2.38, 3.15){\scriptsize \textbf{Contact}}

\put(4.15, 3.7){\rotatebox[origin=c]{-90}{\scriptsize \textbf{Grasp}}}
\put(3.95, 3.7){\rotatebox[origin=c]{-90}{\scriptsize \textbf{Releasing}}}

\put(4.15, 2.35){\rotatebox[origin=c]{-90}{\scriptsize \textbf{Finger}}}
\put(3.95, 2.4){\rotatebox[origin=c]{-90}{\scriptsize \textbf{Grazing}}}

\put(4.2, 0.95){\rotatebox[origin=c]{-90}{\scriptsize \textbf{Palm}}}
\put(3.95, 1.0){\rotatebox[origin=c]{-90}{\scriptsize \textbf{Rolling}}}
\end{picture}
\caption{Three contact interaction modes including grasp-releasing (top), finger-grazing (middle), and palm-rolling (bottom).}
\label{fig:contact_modes} 
\end{figure}
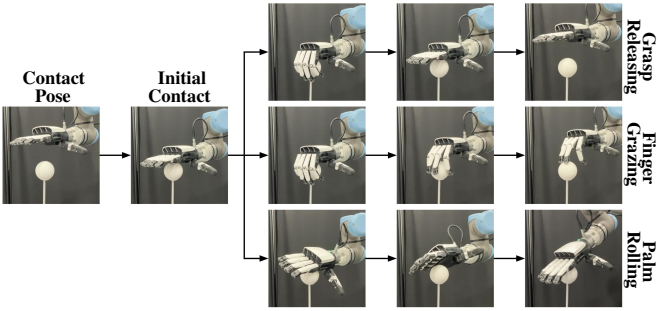

\subsection{Contact Interaction Modes} 
Naive binary grasp-and-release approaches yield limited contact points per interaction. To address this limitation, we consider two alternative contact interaction modes, namely \textit{finger-grazing} and \textit{palm-rolling}, which meaningfully improve tactile sensing efficiency. Our contact strategy implements these interactions in two phases. First, the robot moves its palm-oriented hand toward the target object along a linear path. Upon detecting contact, the robot interacts with the object in one of three interaction modes (see Fig.~\ref{fig:contact_modes}): 
\begin{itemize}
    \item \textit{Grasp-Releasing:} A sequence of power-grasp and releasing motion.
    \item \textit{Finger-Grazing:} Fingers curl inward under torque control while the arm retracts under force control, inducing sliding contacts.
    \item \textit{Palm-Rolling:} Fingers maintain open positions while the arm rotates under force control inducing rolling contacts: (1) ±30 degree bidirectional rotation about the gripper axis, followed by (2) a 30 degree forward pitch about the palm's lateral axis.
\end{itemize}

\section{Experiments}

\subsection{Shape Completion Results on Synthetic Data}
\label{subsec:model_training}

\begin{figure}[t]
\centering 
\includegraphics[width=0.9\columnwidth]{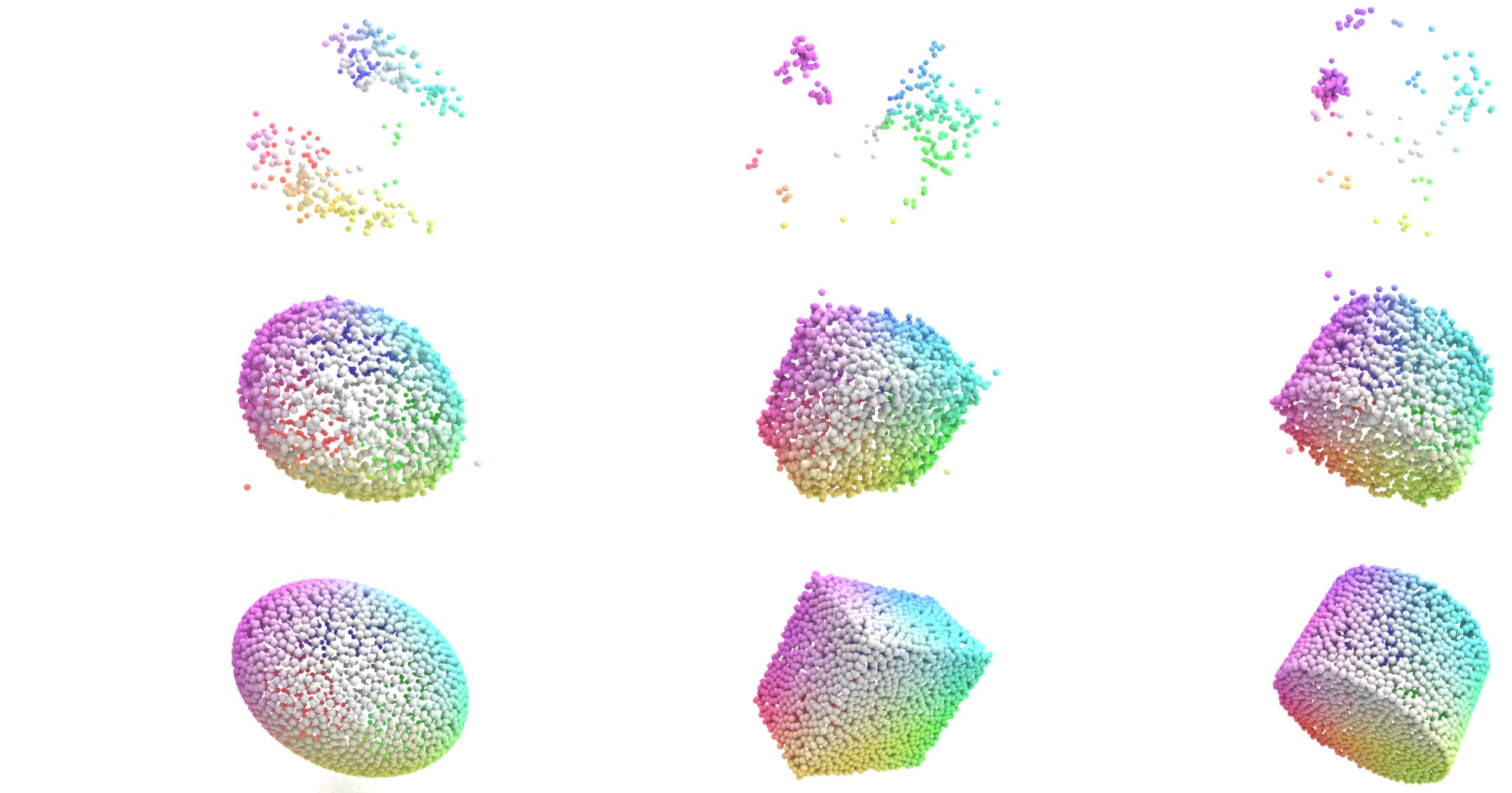}
\setlength{\unitlength}{1cm}
\begin{picture}(0,0)

\put(-7.95, 3.6){{\scriptsize \textbf{Tactile}}}
\put(-7.87, 3.35){{\scriptsize \textbf{Point}}}
\put(-7.93, 3.1){{\scriptsize \textbf{Cloud}}}
\put(-7.93, 2.2){{\scriptsize \textbf{Shape}}}
\put(-8.25, 1.95){{\scriptsize \textbf{Completion}}}
\put(-8.05, 0.65){{\scriptsize \textbf{Ground}}}
\put(-7.95, 0.4){{\scriptsize \textbf{Truth}}}

\end{picture}
\caption{Shape completion results on simulated tactile point clouds.}
\label{fig:synthetic_results} 
\end{figure}
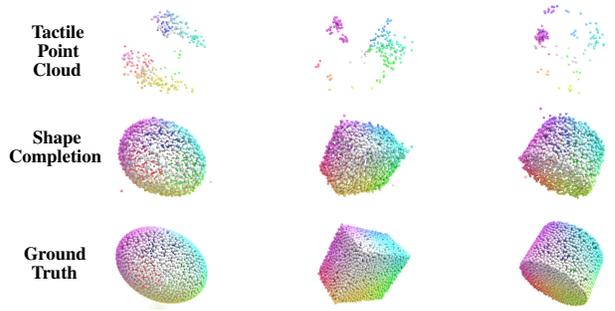

We evaluated our shape completion model on a synthetic dataset comprising three primitive shapes: balls, boxes, and cylinders \cite{papert1966summer}. For each shape category, we generated a dataset that consists of 100 training, 15 validation, and 15 test meshes with randomized aspect ratios. Each mesh underwent 20 simulated contact-grasp interactions to create input-output data sample pairs. To enhance real-world robustness of the model, we augmented the dataset by applying point truncation ranging from 0\% to 90\% in 10\% increments, resulting in 3,000 training samples. During training, we dynamically applied Gaussian noise and random rotations. The model was trained for 2,000 epochs with a batch size of 64 and learning rate of 5e-4. We used a DDIM scheduler with 1,000 training steps and 100 inference steps to balance computational efficiency and reconstruction quality, resulting in an inference time of 2.5 seconds per sample on an NVIDIA GeForce RTX 2080 Super GPU. Fig.~\ref{fig:synthetic_results} shows qualitative results on the synthetic test dataset, demonstrating the model’s ability to accurately reconstruct complete object shapes from partial observations. 

\subsection{Real-World Experiment Setup}
\label{subsec:experiment_setup} 

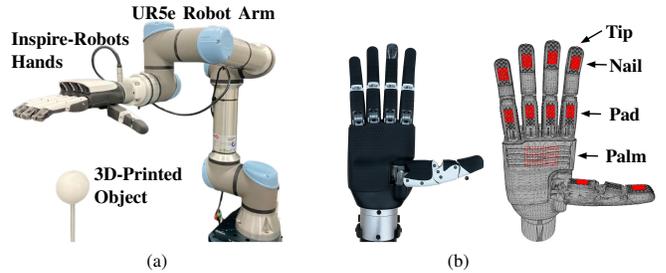
\begin{figure}[t]
\centering 
\includegraphics[width=\columnwidth]{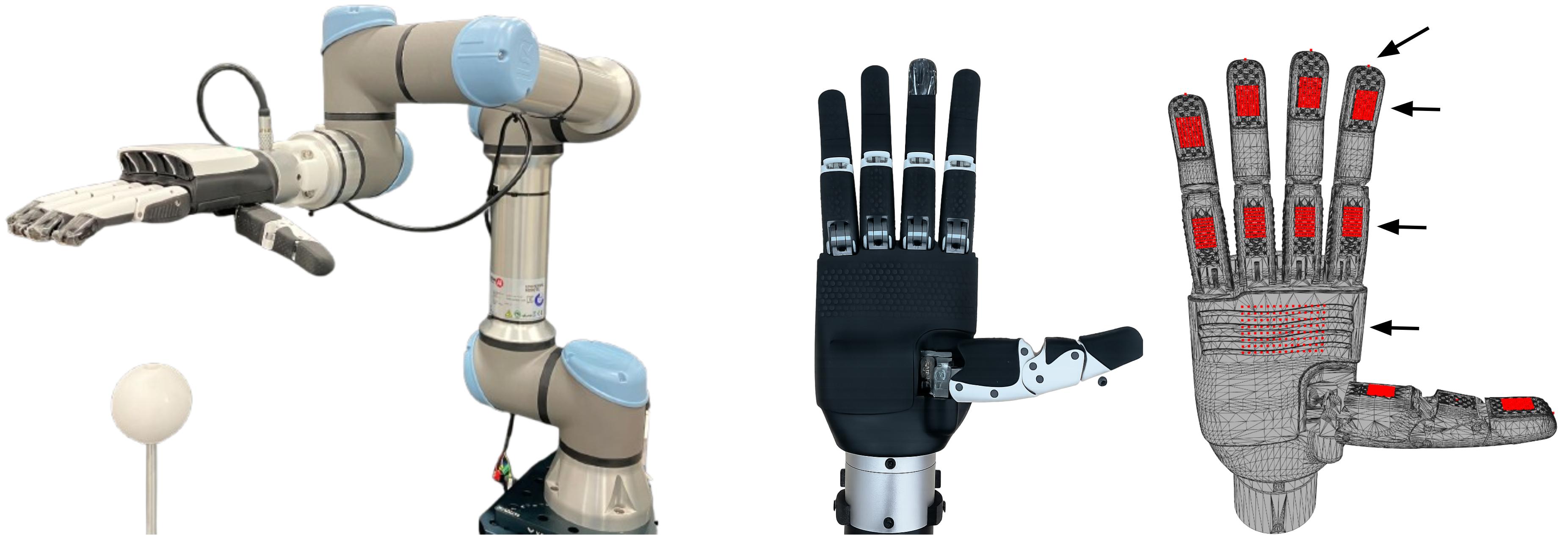}
\setlength{\unitlength}{1cm}
\begin{picture}(0,0)
\put(-2.5, 0.1){\scriptsize (a)}
\put(1.5, 0.1){\scriptsize (b)}
\put(3.6, 3.15){\scriptsize \textbf{Tip}}
\put(3.65, 2.7){\scriptsize \textbf{Nail}}
\put(3.65, 2.05){\scriptsize \textbf{Pad}}
\put(3.6, 1.5){\scriptsize \textbf{Palm}}
\put(-2.7, 3.4){\scriptsize \textbf{UR5e Robot Arm}}
\put(-4.3, 3.05){\scriptsize \textbf{Inspire-Robots}}
\put(-4.3, 2.75){\scriptsize \textbf{Hands}}
\put(-3.2, 1.25){\scriptsize \textbf{3D-Printed}}
\put(-3.2, 0.95){\scriptsize \textbf{Object}}
\end{picture}
\caption{(a) Our system features a UR5e robot arm and the Inspire-Robots Dexterous Hands embedded with tactile sensors. (b) Tactile sensor arrays on the palmar side of the anthropomorphic end-effector is visualized in red.}
\label{fig:experiment_setup} 
\end{figure}

We conducted real-world experiments using the anthropomorphic Inspire-Robots Dexterous Hand mounted on a UR5e robot arm (Fig.~\ref{fig:experiment_setup}(a)). The hand exhibits five fingers with six degrees of freedom (extra DOF for the thumb) and is equipped with tactile sensing capabilities, featuring pressure sensor arrays distributed across multiple contact surfaces (see Fig.~\ref{fig:experiment_setup}(b)): 3×3 arrays on fingertips, 12×8 on nails, 10×8 on finger pads, and 8×14 on the palm. While our implementation utilizes this specific hardware configuration, our approach is designed to be compatible with any tactile gripper capable of generating contact point cloud data.

\begin{figure*}[t]
\centering 
\includegraphics[width=0.9\textwidth]{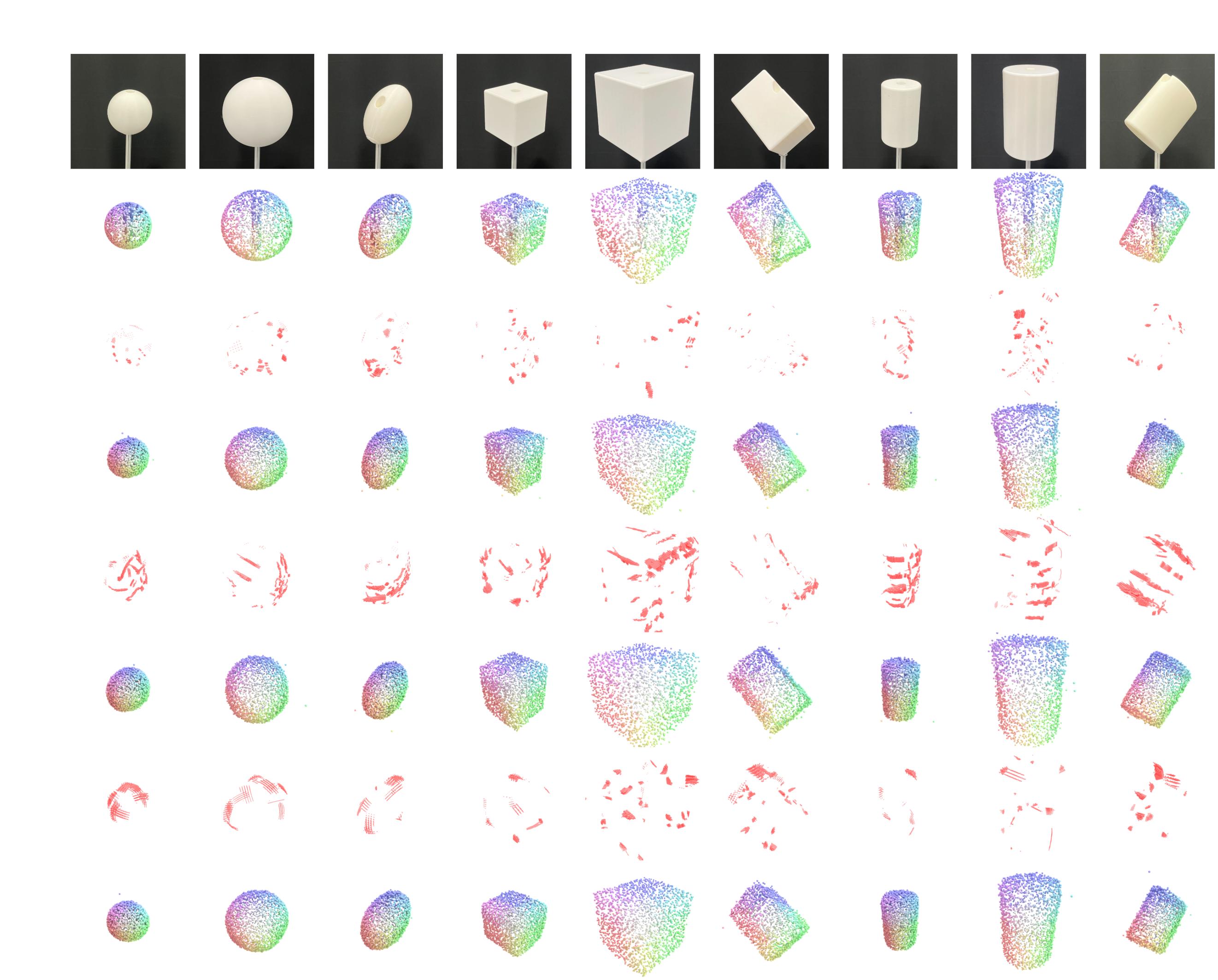}
\setlength{\unitlength}{1cm}
\begin{picture}(0,0)
\put(-13.15, 12.55){\small \textbf{Ball}}
\put(-14.85, 12.25){\scriptsize \textbf{Small}}
\put(-13.05, 12.25){\scriptsize \textbf{Big}}
\put(-11.45, 12.25){\scriptsize \textbf{D\&R\textsuperscript{\dag}}}

\put(-8.1, 12.55){\small \textbf{Box}}
\put(-9.8, 12.25){\scriptsize \textbf{Small}}
\put(-8.0, 12.25){\scriptsize \textbf{Big}}
\put(-6.4, 12.25){\scriptsize \textbf{D\&R\textsuperscript{\dag}}}

\put(-3.4, 12.55){\small \textbf{Cylinder}}
\put(-4.75, 12.25){\scriptsize \textbf{Small}}
\put(-2.95, 12.25){\scriptsize \textbf{Big}}
\put(-1.35, 12.25){\scriptsize \textbf{D\&R\textsuperscript{\dag}}}

\put(-15.75, 11.35){\rotatebox[origin=c]{90}{\scriptsize \textbf{Object}}}
\put(-15.75, 9.8){\rotatebox[origin=c]{90}{\scriptsize \textbf{Ground}}}
\put(-15.45, 9.8){\rotatebox[origin=c]{90}{\scriptsize \textbf{Truth}}}

\put(-16.25, 10.5){\rotatebox[origin=c]{90}{\small \textbf{Reference}}}

\put(-16.25, 7.45){\rotatebox[origin=c]{90}{\small \textbf{Grasp-Releasing}}}

\put(-15.75, 8.25){\rotatebox[origin=c]{90}{\scriptsize \textbf{Tactile}}}
\put(-15.45, 8.25){\rotatebox[origin=c]{90}{\scriptsize \textbf{Input}}}
\put(-15.75, 6.8){\rotatebox[origin=c]{90}{\scriptsize \textbf{Shape}}}
\put(-15.45, 6.8){\rotatebox[origin=c]{90}{\scriptsize \textbf{Output}}}

\put(-16.25, 4.45){\rotatebox[origin=c]{90}{\small \textbf{Finger-Grazing}}}

\put(-15.75, 5.25){\rotatebox[origin=c]{90}{\scriptsize \textbf{Tactile}}}
\put(-15.45, 5.25){\rotatebox[origin=c]{90}{\scriptsize \textbf{Input}}}
\put(-15.75, 3.75){\rotatebox[origin=c]{90}{\scriptsize \textbf{Shape}}}
\put(-15.45, 3.75){\rotatebox[origin=c]{90}{\scriptsize \textbf{Output}}}

\put(-16.25, 1.4){\rotatebox[origin=c]{90}{\small \textbf{Palm-Rolling}}}

\put(-15.75, 2.25){\rotatebox[origin=c]{90}{\scriptsize \textbf{Tactile}}}
\put(-15.45, 2.25){\rotatebox[origin=c]{90}{\scriptsize \textbf{Input}}}
\put(-15.75, 0.7){\rotatebox[origin=c]{90}{\scriptsize \textbf{Shape}}}
\put(-15.45, 0.7){\rotatebox[origin=c]{90}{\scriptsize \textbf{Output}}}

\end{picture}
\caption{Qualitative results of shape reconstruction on real-world tactile data. Tactile measurements are obtained through information-theoretic exploration using one of three contact interaction modes. We tested our framework on nine objects fixed relative to the robot base. \textbf{\textsuperscript{\dag}D\&R}: Deformed \& Rotated.}
\label{fig:real_world_results} 
\end{figure*}
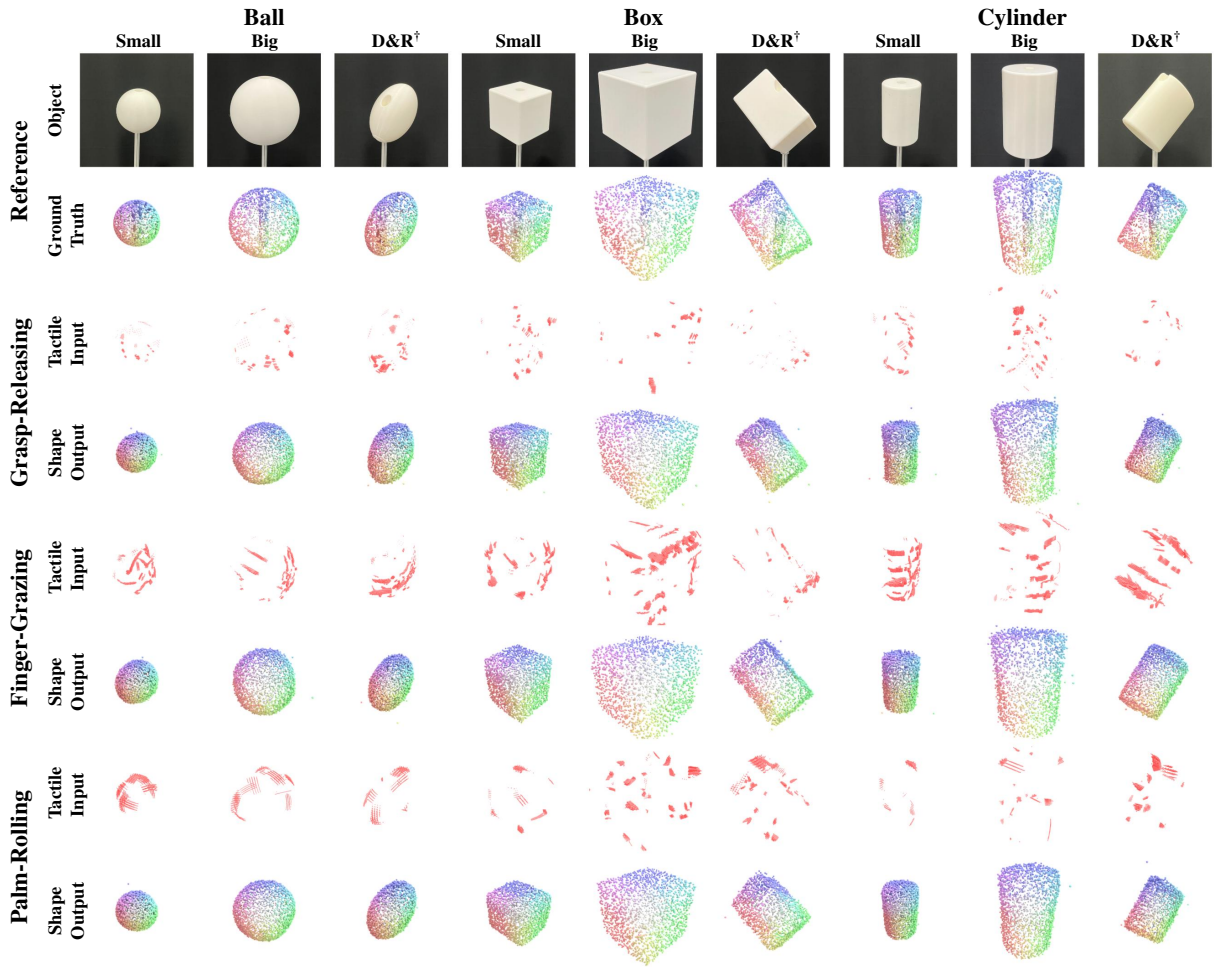

\begin{figure*}[t]
\centering 
\includegraphics[width=0.8\textwidth]{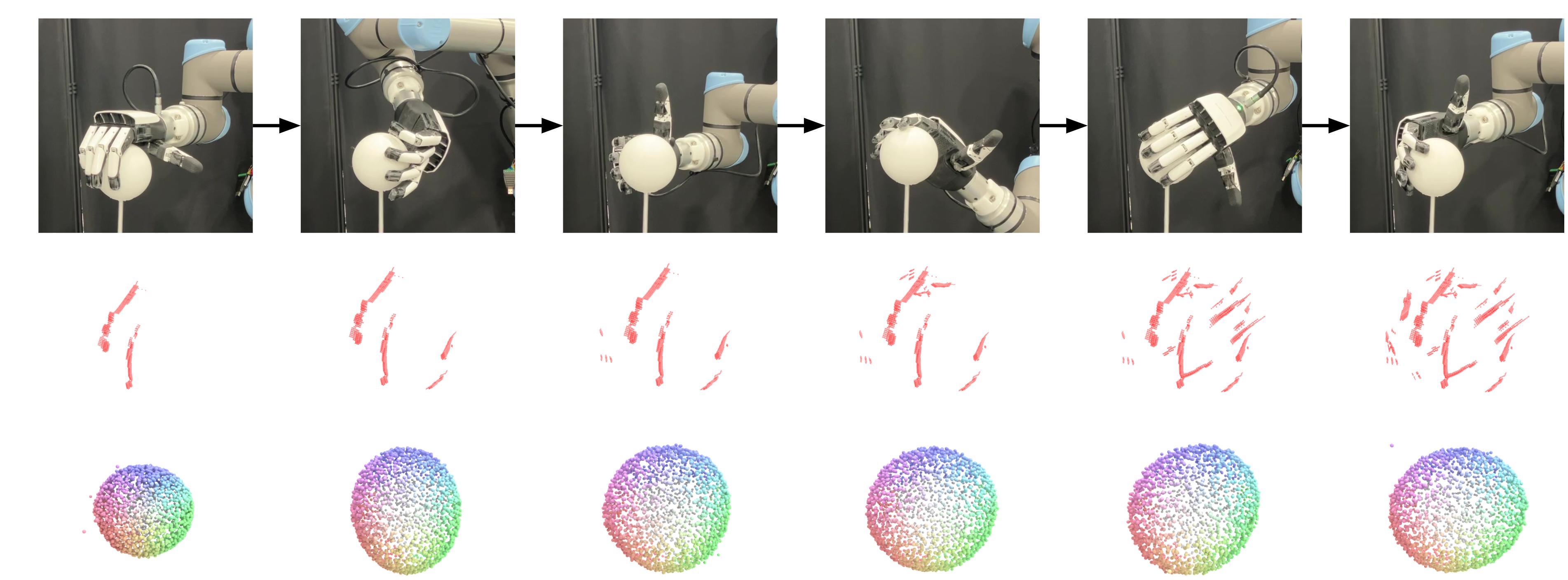}
\setlength{\unitlength}{1cm}
\begin{picture}(0,0)
\put(-13.4, 5.2){\scriptsize \textbf{Iter 1}}
\put(-11.05, 5.2){\scriptsize \textbf{Iter 2}}
\put(-8.7, 5.2){\scriptsize \textbf{Iter 3}}
\put(-6.3, 5.2){\scriptsize \textbf{Iter 4}}
\put(-3.95, 5.2){\scriptsize \textbf{Iter 5}}
\put(-1.5, 5.2){\scriptsize \textbf{Iter 6}}

\put(-14.75, 4.05){\rotatebox[origin=c]{90}{\scriptsize \textbf{Robot}}}
\put(-14.45, 4.05){\rotatebox[origin=c]{90}{\scriptsize \textbf{Exploration}}}
\put(-14.75, 2.15){\rotatebox[origin=c]{90}{\scriptsize \textbf{Tactile}}}
\put(-14.45, 2.15){\rotatebox[origin=c]{90}{\scriptsize \textbf{Measurement}}}
\put(-14.75, 0.55){\rotatebox[origin=c]{90}{\scriptsize \textbf{Shape}}}
\put(-14.45, 0.55){\rotatebox[origin=c]{90}{\scriptsize \textbf{Prediction}}}
\end{picture}
\caption{Sequential robotic tactile exploration of a 3D-printed ball using finger-grazing motions. (Top row) Contact locations from each exploration step. (Middle row) Accumulated tactile point cloud data. (Bottom row) Progressive shape reconstruction results. See supplementary video for detailed visualizations.
}
\label{fig:iteration} 
\end{figure*}
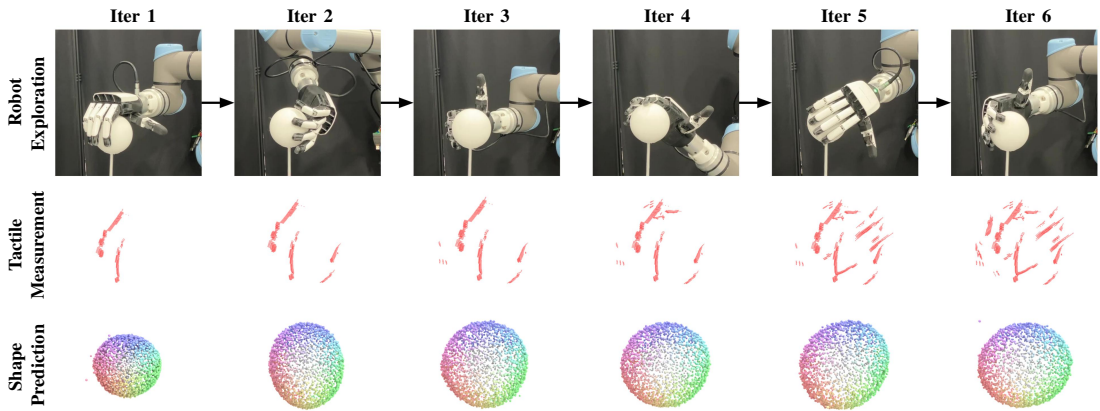

We tested our approach on nine 3D-printed objects comprised of three primitive shapes. For each shape category, we included three variations: a standard geometric form (perfect sphere, cube and cylinder), a scaled variant 1.5 times larger, and a deformed variant with modified aspect ratios and an orientation offset of 45-degrees (see Fig.~\ref{fig:real_world_results} and Table~\ref{table:quant_results}). Each object was fixed relative to the robot base and was tested across 5 different initial robot poses per contact mode. To ensure consistent initialization, we manually positioned the robot such that linear movement along the palm-axis would result in object contact during the first interaction. Episodes were limited to a maximum of 20 contact interactions or until convergence criteria was met (discussed in Sec.~\ref{subsec:exploration}). This experimental design yielded 135 total trials (3 contact modes × 9 objects × 5 trials). The shape completion model, trained exclusively on synthetic data as described in Sec.~\ref{subsec:model_training}, was applied directly without additional fine-tuning, with each exploration episode conditioned only on the guaranteed contact initialization and the pre-trained model.

\begin{figure*}[t]
\centering 
\includegraphics[width=0.9\textwidth]{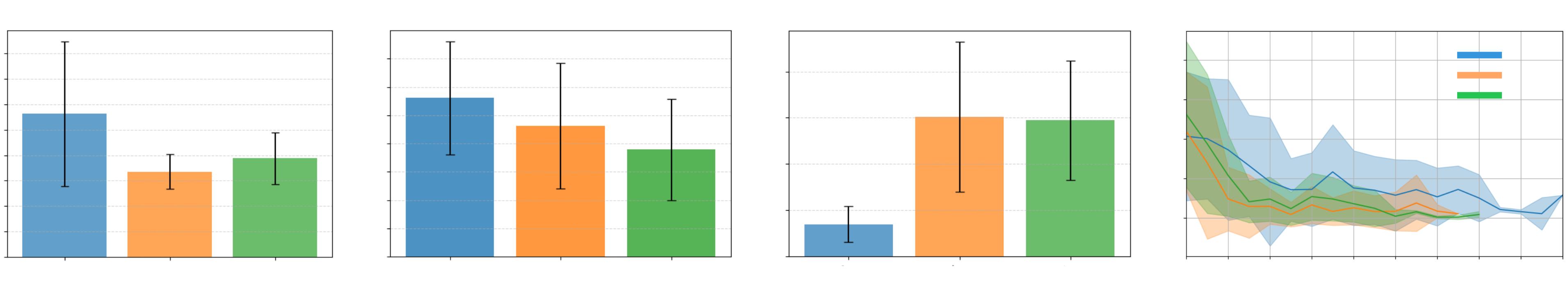}
\setlength{\unitlength}{1cm}
\begin{picture}(0,0)
\put(-14.65, -0.3){\footnotesize (a)}
\put(-10.65, -0.3){\footnotesize (b)}
\put(-6.6, -0.3){\footnotesize (c)}
\put(-2.3, -0.3){\footnotesize (d)}
\put(-15.75, 0.15){\scriptsize \textbf{GR}}
\put(-14.65, 0.15){\scriptsize \textbf{FG}}
\put(-13.55, 0.15){\scriptsize \textbf{PR}}

\put(-11.85, 0.15){\scriptsize \textbf{GR}}
\put(-10.7, 0.15){\scriptsize \textbf{FG}}
\put(-9.55, 0.15){\scriptsize \textbf{PR}}

\put(-7.75, 0.15){\scriptsize \textbf{GR}}
\put(-6.6, 0.15){\scriptsize \textbf{FG}}
\put(-5.45, 0.15){\scriptsize \textbf{PR}}

\put(-0.87, 2.42){\scriptsize \textbf{GR}}
\put(-0.85, 2.20){\scriptsize \textbf{FG}}
\put(-0.85, 1.98){\scriptsize \textbf{PR}}

\put(-16.4, 2.8){\scriptsize \textbf{Reconstruction Chamfer Distance}}
\put(-16.65, 1.5){\rotatebox[origin=c]{90}{\tiny \textbf{Chamfer Distance}}}
\put(-16.4, 2.45){\rotatebox[origin=c]{90}{\tiny 0.16}}
\put(-16.4, 1.9){\rotatebox[origin=c]{90}{\tiny 0.12}}
\put(-16.4, 1.35){\rotatebox[origin=c]{90}{\tiny 0.08}}
\put(-16.4, 0.85){\rotatebox[origin=c]{90}{\tiny 0.04}}
\put(-16.4, 0.35){\rotatebox[origin=c]{90}{\tiny 0}}
\put(-16.4, 0.35){\rotatebox[origin=c]{90}{\tiny 0}}

\put(-12.35, 2.8){\scriptsize \textbf{\# Interactions until Convergence}}
\put(-12.7, 1.4){\rotatebox[origin=c]{90}{\tiny \textbf{\# Interactions}}}
\put(-12.5, 2.35){\rotatebox[origin=c]{90}{\tiny 14}}
\put(-12.5, 2.05){\rotatebox[origin=c]{90}{\tiny 12}}
\put(-12.5, 1.75){\rotatebox[origin=c]{90}{\tiny 10}}
\put(-12.5, 1.5){\rotatebox[origin=c]{90}{\tiny 8}}
\put(-12.5, 1.2){\rotatebox[origin=c]{90}{\tiny 6}}
\put(-12.5, 0.9){\rotatebox[origin=c]{90}{\tiny 4}}
\put(-12.5, 0.62){\rotatebox[origin=c]{90}{\tiny 2}}
\put(-12.5, 0.35){\rotatebox[origin=c]{90}{\tiny 0}}

\put(-8.2, 2.8){\scriptsize \textbf{Contact Volume per Interaction}}
\put(-8.7, 1.4){\rotatebox[origin=c]{90}{\tiny \textbf{Contact Volume ($\text{mm}^3$)}}}
\put(-8.4, 2.25){\rotatebox[origin=c]{90}{\tiny 400}}
\put(-8.4, 1.78){\rotatebox[origin=c]{90}{\tiny 300}}
\put(-8.4, 1.3){\rotatebox[origin=c]{90}{\tiny 200}}
\put(-8.4, 0.8){\rotatebox[origin=c]{90}{\tiny 100}}
\put(-8.4, 0.35){\rotatebox[origin=c]{90}{\tiny 0}}

\put(-3.85, 2.8){\scriptsize \textbf{Chamfer Distance Progression}}

\put(-4.55, 1.4){\rotatebox[origin=c]{90}{\tiny \textbf{Chamfer Distance}}}
\put(-4.33, 1.95){\rotatebox[origin=c]{90}{\tiny 0.2}}
\put(-4.33, 1.15){\rotatebox[origin=c]{90}{\tiny 0.1}}
\put(-4.33, 0.35){\rotatebox[origin=c]{90}{\tiny 0}}

\put(-3.75, 0.2){\tiny 2}
\put(-3.3, 0.2){\tiny 4}
\put(-2.9, 0.2){\tiny 6}
\put(-2.45, 0.2){\tiny 8}
\put(-2.1, 0.2){\tiny 10}
\put(-1.67, 0.2){\tiny 12}
\put(-1.25, 0.2){\tiny 14}
\put(-0.8, 0.2){\tiny 16}
\put(-0.4, 0.2){\tiny 18}
\put(-2.75, 0.0){\tiny \textbf{\# Interactions}}

\end{picture}
\vspace{8pt}
\caption{
Key metrics across different contact modes -- \textbf{GR:} Grasp-Releasing / \textbf{FG:} Finger-Grazing / \textbf{PR:} Palm-Rolling.
(a) \textbf{Reconstruction Chamfer Distance:} Average Chamfer distance for shape reconstruction at convergence.
(b) \textbf{Number of Interactions Until Convergence:} Average number of interactions required to achieve shape reconstruction convergence.
(c) \textbf{Contact Volume per Interaction:} Average volume of tactile data collected during each interaction.
(d) \textbf{Chamfer Distance Progression:} Illustrates how reconstruction Chamfer distance decreases over successive interactions.
} 
\label{fig:graphs} 
\end{figure*}

\begin{table*}[htbp]
\captionsetup{font=small} 
\caption{Quantitative Results for Real-World Tactile Shape Reconstruction}
\centering
\setlength{\tabcolsep}{8pt} 
\renewcommand{\arraystretch}{1.3} 
\begin{tabularx}{\textwidth}{@{\extracolsep{\fill}}cccccccccccc@{}}
\hline\hline
\textbf{Shape} &  \multirow{2}{*}{{\shortstack[c]{\textbf{Variation}}}} & \textbf{Dimension} &  \multicolumn{3}{c}{\textbf{Chamfer Distance}} & \multicolumn{3}{c}{\textbf{\# Interactions to Converge}} & \multicolumn{3}{c}{\textbf{Contact Volume (mm$^3$)}} \\ 
\textbf{Category} &  & \textbf{(mm)} & \textbf{GR} & \textbf{FG} & \textbf{PR}  & \textbf{GR} & \textbf{FG} & \textbf{PR} & \textbf{GR} & \textbf{FG} & \textbf{PR} \\
\hline 
\multirow{3}{*}{{\shortstack[c]{Ball}}} & Small & $r_x=r_y=r_z=30$ & 0.079 & \textbf{0.046} & 0.048 & 6.8 & 5.2 & \textbf{3.2} & 52 & 263 & \textbf{472}\\
                                        & Big & $r_x=r_y=r_z=45$ & 0.095 & \textbf{0.073} & 0.088 & 10.0 & \textbf{5.4} & 7.0 & 114 & 174 & \textbf{404}\\
                                        & D\&R\textsuperscript{\dag} & $r_x=r_y=30, r_z=45$ & \textbf{0.047} & 0.048 & 0.060 & 11.0 & 6.2 & \textbf{4.8} & 89 & \textbf{377} & 319\\
\hline 
\multirow{3}{*}{{\shortstack[c]{Box}}} & Small & $w=h=l=60$  & 0.102 & 0.060 & \textbf{0.042} & 13.4 & 11.6 & \textbf{7.2} & 65 & \textbf{417} & 137\\
                                       & Big   & $w=h=l=90$  & 0.099 & \textbf{0.090} & 0.095 & 11.8 & \textbf{10.6} & 11.4& 77 & 238 & \textbf{292}\\
                                       & D\&R\textsuperscript{\dag}  & $w=h=60, l=90$ & 0.243 & \textbf{0.062} & 0.100 & 13.6 & 16.6 & \textbf{10.0} & 39 & 234 & \textbf{362}\\
\hline 
\multirow{3}{*}{{\shortstack[c]{Cylinder}}} & Small & $r_x=r_y=25, h=80$ &  0.084 & \textbf{0.050} & 0.074 & 9.6 & \textbf{4.8} & \textbf{4.8} & 50 & \textbf{384} & 189\\
                                       & Big   & $r_x=r_y=37.5, h=120$ &  0.191 & \textbf{0.114} & 0.125 & 12.6 & 12.4 & \textbf{8.4} & 80 & \textbf{242} & 200\\
                                       & D\&R\textsuperscript{\dag}  & $r_x=25, r_y=37.5, h=80$ & 0.070 & \textbf{0.051} & 0.059 & 12.6 & \textbf{10.4} & 11.4 & 65 & \textbf{426} & 283\\
\hline\hline
\end{tabularx}
\captionsetup{justification=centering}
\begin{minipage}{\textwidth}
\vspace{1mm}
\footnotesize 
*All entries are averaged over 5 trials, initialized from different end-effector poses. \textbf{GR}: Grasp-Releasing / \textbf{FG}: Finger-Grazing / \textbf{PR}: Palm-Rolling \\
\textsuperscript{\dag}\textbf{D\&R}: Deformed \& Rotated (45$^\circ$)
\end{minipage}
\label{table:quant_results}
\end{table*}

\subsection{Shape Completion on Real-World Tactile Data}

Fig.~\ref{fig:real_world_results} presents qualitative examples of object shape reconstruction through tactile exploration using the three interaction modes (grasp-releasing, finger-grazing, and palm-rolling). Our results demonstrate successful sim-to-real transfer, with the shape completion model effectively generalizing to real-world data. This transferability can be attributed to both the relatively low sim-to-real gap of point cloud representations (compared to RGB modalities) and the use of comprehensive augmentation and normalization strategies. The results further show that all three modes were capable of reconstructing the object when guided by the information-theoretic exploration policy described in Sec.~\ref{subsec:exploration} and illustrated in Fig.~\ref{fig:iteration} (see supplementary video).

\subsection{Quantitative Results \& Discussion}
Despite this shared capability, the reconstruction accuracy and the number of interactions required for convergence varied across contact modes. Detailed quantitative results are reported in Table~\ref{table:quant_results}. The mean Chamfer distances at convergence were 0.114, 0.067, and 0.077 (Fig.~\ref{fig:graphs}(a)), with corresponding mean interaction counts of 11.2, 9.2, and 7.6 (Fig.~\ref{fig:graphs}(b)) for grasp-releasing, finger-grazing, and palm-rolling, respectively. We further quantified the tactile information acquired per interaction by voxelizing the accumulated tactile point clouds at a resolution of 1 mm. The resulting mean contact volumes were 70, 302, and 294 mm$^3$ per interaction for the three modes, respectively (Fig.~\ref{fig:graphs}(c)). 

Our experimental results demonstrate distinct advantages and trade-offs among the three contact interaction modes. The dynamic nature of sliding and rolling interactions proved particularly effective at gathering rich tactile information. While grasp-releasing provides discrete contact points, both finger-grazing and palm-rolling enable continuous surface sampling through sweeping motions, resulting in approximately four times more tactile data acquisition per interaction in terms of contact volume. This enhanced data density translated to improved reconstruction accuracy, with finger-grazing and palm-rolling achieving 70\% and 48\% improved accuracy respectively, while requiring 22\% and 47\% fewer contact interactions compared to grasp-releasing. This efficiency is evident in Fig.~\ref{fig:graphs}(d), which illustrates the faster convergence rates of finger-grazing and palm-rolling in terms of Chamfer distance reduction over episode progression.

However, each interaction mode presents distinct operational considerations influenced by both hardware constraints and control complexity. Grasp-releasing, while yielding the least tactile data per interaction, offers the advantages of minimal execution time and reduced sensor noise due to the absence of motion-induced latency issues. Finger-grazing achieves efficient data acquisition but requires precise coordination between arm and hand motions, introducing potential latency challenges in real-world implementation. Palm-rolling, despite requiring longer execution times due to extended arm rotations, minimizes finger motion-related noise and proves particularly effective for curved surfaces. However, its performance is notably constrained when exploring flat surfaces (in box-like objects), primarily due to the limited tactile surface area and the concave geometry of the palmar side of the hand. These findings highlight the importance of considering both the geometric properties of the tactile sensor surfaces and the target object characteristics when selecting an appropriate interaction strategy.

\section{Conclusion}
In this work, we investigated the effectiveness of different contact modes for tactile data acquisition for the task of object reconstruction using an anthropomorphic hand equipped with tactile sensors. A diffusion-based shape completion model, trained on simulated data, was employed to reconstruct complete object geometry from partial tactile point clouds. Real-world experiments, guided by an information-theoretic exploration framework, compare three interaction modes: grasp-releasing, finger-grazing, and palm-rolling. The results show that dynamic contact modes achieve a reconstruction accuracy improvement of 55\% while requiring 34\% fewer interactions than basic grasping, underscoring the value of diverse contact strategies for efficient tactile exploration. Future directions include adaptive contact mode selection informed by local geometry inference and real-time tactile feedback.

\bibliographystyle{ieeetr}
\bibliography{main}

\end{document}